\documentclass[letterpaper, 10 pt, conference]{ieeeconf}

\usepackage{graphicx}
\usepackage{balance}
\usepackage{comment}
\usepackage{cite}
\usepackage{amssymb}
\usepackage[tight,footnotesize]{subfigure}
\usepackage[active]{srcltx}
\usepackage{amsmath}

\graphicspath{{./figs/}}
\renewcommand{\baselinestretch}{0.95}
\usepackage{eurosym}
\usepackage[plain]{algorithm}
\usepackage{algorithmic}
\usepackage{multicol}
\usepackage{dsfont}
\usepackage{amsfonts}

\usepackage{cases}
\usepackage{xcolor}

\IEEEoverridecommandlockouts
\begin{document}

\title{Unscented Optimal Control for 3D Coverage Planning \\with an Autonomous UAV Agent}

\author{Savvas~Papaioannou,~Panayiotis~Kolios,~Theocharis~Theocharides,\\~Christos~G.~Panayiotou~ and ~Marios~M.~Polycarpou
\thanks{The authors are with the KIOS Research and Innovation Centre of Excellence (KIOS CoE) and the Department of Electrical and Computer Engineering, University of Cyprus, Nicosia, 1678, Cyprus. {\tt\small \{papaioannou.savvas, pkolios, ttheocharides, christosp, mpolycar\}@ucy.ac.cy}}
}

\maketitle

\begin{abstract}
We propose a novel probabilistically robust controller for the guidance of an unmanned aerial vehicle (UAV) in coverage planning missions, which can simultaneously optimize both the UAV's motion, and camera control inputs for the 3D coverage of a given object of interest. Specifically, the coverage planning problem is formulated in this work as an optimal control problem with logical constraints to enable the UAV agent to jointly: a) select a series of discrete camera field-of-view states which satisfy a set of coverage constraints, and b) optimize its motion control inputs according to a specified mission objective. We show how this hybrid optimal control problem can be solved with standard optimization tools by converting the logical expressions in the constraints into equality/inequality constraints involving only continuous variables. Finally, probabilistic robustness is achieved by integrating the unscented transformation to the proposed controller, thus enabling the design of robust open-loop coverage plans which take into account the future posterior distribution of the UAV's state inside the planning horizon.
\end{abstract}

\section{Introduction} \label{sec:Introduction}

The interest in unmanned aerial vehicles (UAVs), and their utilization in various applications domains such as security \cite{PapaioannouJ1,Wu2019,PapaioannouJ2,PapaioannouTMC2022}, emergency response \cite{Papaioannou2021a,Papaioannou2021b,Li2021energy}, monitoring/searching \cite{skjong2015autonomous,Papaioannou2020,PapaioannouTMC,Kopfstedt2008control,PapaioannouICUAS2019}, air-traffic management \cite{shrestha20216g,vitale2022}, and automated infrastructure inspection \cite{Shukla2016,Luo2017,Stoican2019,PapaioannouICUAS2022} has peaked in the last years.
The majority of the application domains mentioned above require some form of coverage planning\cite{cabreira2019survey}, which in general requires the UAV to be able to plan an optimal trajectory for the coverage of a specific set of points or objects inside an area of interest. Designing fully autonomous coverage missions with UAVs however requires robust planning methodologies. More specifically, the utilization of autonomous UAVs in challenging application domains such as the ones discussed above is directly correlated with their ability to cope with, and minimize the effects of uncertainty caused by external disturbances, modelling mismatch, and sensor imperfections. 
Typically, closed-loop control \cite{Anderson2013,franklin2002feedback} is utilized in such scenarios to mitigate the effects of the uncertainty, and generate reliable trajectories. These approaches rely on the availability of accurate sensor readings (e.g., availability of GPS signals), that can be used for updating the control inputs to compensate for the disturbances. However, there are situations where sensor measurements cannot be utilized either because they are contaminated with noise, or because they are completely unavailable. In such scenarios, close-loop control cannot be used, and therefore it is desirable to compute robust open-loop trajectories which do not rely on sensor readings. 

In this work, we combine the unscented transformation (UT) \cite{UT1} with optimal control, to design a probabilistically robust open-loop control law which can guide an autonomous UAV agent to cover in 3D the surface area of an object of interest in the presence of uncertainty caused by random disturbances on the UAV's dynamics. In principle, as discussed in detail in Sec. \ref{sec:approach}, the sigma-points of the UT act as controllable particles guided by the same open-loop controller, and are used as a surrogate for the characterization of the posterior distribution of the UAV's state. As a result, the future probability distribution of the agent's state can be guided inside a finite planning horizon to meet a certain set of probabilistic mission constraints. The proposed unscented optimal control problem is then integrated with a set of logical constraints, which are responsible for the simultaneous optimization of the UAV's motion and camera control inputs. In particular, these logical constraints enable the guidance of the UAV through a series of waypoints (the waypoint traversal order is determined during the optimization), and subsequently the selection of a set of discrete camera field-of-view (FOV) states which result in the coverage of the required surface area. The contributions of this work can be summarized as follows:

\begin{itemize}
    \item We propose a novel coverage controller for an autonomous UAV agent which generates probabilistically robust coverage plans for 3D objects. We show how probability constraints can be translated into equivalent deterministic constraints which, via the unscented transformation, take into account the probability distribution of the agent's state. 
    \item We formulate the coverage planning problem as an optimal control problem with logical constraints to allow the simultaneous optimization of the UAV's motion and camera control inputs. Subsequently, we show that these logical constraints can be realized through equality and inequality expressions involving only continuous variables, and therefore the problem can be solved using standard off-the-shelf optimization tools.
\end{itemize}

The rest of the paper is organized as follows. Section~\ref{sec:Related_Work} summarizes the related work on coverage planning. Section \ref{sec:Preliminaries} discusses our modelling assumptions, and Section \ref{sec:problem} formulates the problem tackled in this work. Then, Section \ref{sec:approach} discusses the details of the proposed approach, and Section \ref{sec:Evaluation} evaluates the proposed approach. Finally, Section \ref{sec:conclusion} concludes the paper and discusses future work.

\section{Related Work}\label{sec:Related_Work}

A good starting point for the problem of coverage path planning with ground robots is the survey paper in \cite{Galceran2013}. Another survey paper emphasizing the utilization of UAVs in coverage path planning can be found in \cite{Cabreira2019}, and more recently a comprehensive review of the different coverage planning techniques can be found in \cite{Tan2021}. 

Related to the proposed approach, the work in \cite{Xu2011} introduces an automated terrain coverage algorithm using a fixed-wing UAV. The authors use the Boustrophedon cellular decomposition (BCD) method to partition the free space into non-overlapping cells, and then find an Eulerian circuit through all connected cells by solving a linear program. A coverage technique with multiple agents also based on BCD is illustrated in \cite{karapetyan2017efficient} for planar environments. The problem of terrain coverage with a UAV, is also investigated in \cite{Wang2017} under photogrammetric constraints, whereas the approach presented in \cite{Maza2007} tackles the terrain coverage problem for rectilinear environments with multiple UAV agents. The 3D terrain coverage problem with a team of UAVs is investigated in \cite{Renzaglia2011}, where the objective is to find the deployment of UAVs which achieves maximal coverage, while avoiding collisions with the obstacles in the environment. 

The problem of area coverage planning with uncertain robot poses is investigated in \cite{paull2014area}. The authors propose the use of a probabilistic coverage map which is recursively updated using sensor readings, and which in turn can be used to plan coverage paths that take into account the uncertainty on the robot's pose. Moreover, the work in  \cite{jung2009efficient} proposes a coverage planning algorithm for the exploration of the oceanic terrain using multiple underwater vehicles which experience sea current disturbances, and the survey paper in \cite{Dadkhah2012survey} discuses in more detail the problem of motion planning and guidance in the presence of uncertainty.

The coverage of a 3D object of interest with a UAV is posed in \cite{Jing2016} as a view-planning problem. In particular, the authors in \cite{Jing2016} propose a sampling based technique which generates candidate viewpoints of the object, and then combinatorial optimization is applied to find the most suitable set of viewpoints for coverage. The work in \cite{Dornhege2013} also formulates the 3D coverage planning problem as a view-planning problem, and proposes several greedy algorithms on how to generate candidate viewpoints that maximally cover the object of interest. However, the techniques based on view-planning usually do not generate continuous coverage trajectories based on the underlying robot's dynamics, rather they produce a series of discrete robot states. 

A 3D coverage planning approach which extends the rapidly-exploring random tree method is proposed in \cite{Englot2012}, with the objective to find a feasible sampling-based path which covers the object of interest. More recently, the work in \cite{Papaioannou2023} proposed mixed-integer programming technique to tackle the problem of 3D coverage planning of a cuboid-like structures with autonomous UAV agents, whereas  in \cite{Theile2020} and \cite{Sanna2021} the coverage problem is formulated as a learning problem, and solved using deep reinforcement learning and imitation learning respectively. Finally, in \cite{PapaioannouTAES2022} and \cite{PapaioannouCDC2022} the coverage planning problem is formulated as an optimal control problem with visibility constraints, and solved using mixed integer optimization, and reinforcement learning respectively.

Despite the progress in this domain, more work is required until coverage planning can be utilized in fully autonomous flight missions. The majority of the related work: a) only considers 2D planar environments \cite{Galceran2013}, or focuses on the task of terrain coverage \cite{Wang2017,Renzaglia2011,Maza2007}, and not in the task of covering 3D objects of interest, and b) does not jointly optimizes the UAV's motion and camera control inputs \cite{Xu2011,PapaioannouICUAS2022,Englot2012}. Finally, the number of robust coverage planning approaches in the literature is very limited \cite{Cabreira2019,Tan2021}.

\begin{figure}
	\centering
	\includegraphics[scale=0.6]{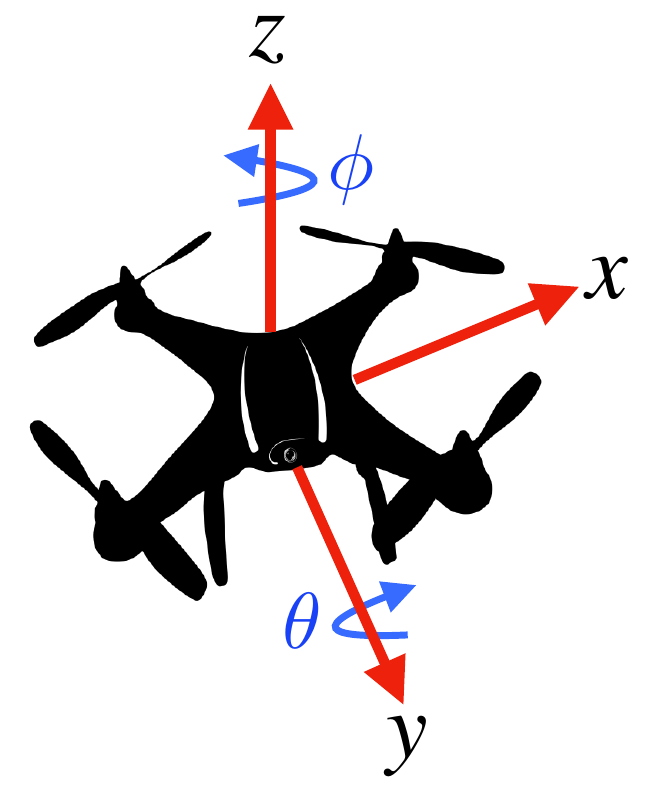}
	\caption{The agent's state $\boldsymbol{x}_t = [x_t,y_t,z_t,\theta_t,\phi_t]$ is composed of position i.e., $\boldsymbol{x}^p_t = [x_t,y_t,z_t]$, and orientation i.e., $\boldsymbol{x}^o_t = [\theta_t,\phi_t]$ components, with motion dynamics expressed by the state-space model shown in Eq. \eqref{eq:dynamics}.}
	\label{fig:fig1}
	\vspace{-5mm}
\end{figure}

\section{Preliminaries} \label{sec:Preliminaries}

\subsection{UAV Dynamical Model} \label{ssec:agent_dynamics}
An autonomous UAV agent operates inside a bounded 3D environment $\mathcal{E} \subset \mathbb{R}^3$ with motion dynamics described by the following discrete-time non-linear state-space model:
\begin{subequations} \label{eq:dynamics}
\begin{align} 
& x_{t+1} = x_t + \Delta T \left(u^v_t + \omega^v_t\right) \cos(\phi_t)\sin(\theta_t), \\
& y_{t+1} = y_t + \Delta T \left(u^v_t + \omega^v_t\right) \sin(\phi_t)\cos(\theta_t), \\
& z_{t+1} = z_t + \Delta T \left(u^v_t + \omega^v_t\right) \sin(\theta_t), \\
& \theta_{t+1} = \theta_t + \Delta T \left(u^\theta_t + \omega^\theta_t\right), \\
& \phi_{t+1} = \phi_t + \Delta T (u^\phi_t + \omega^\phi_t),
\end{align}
\end{subequations}

\noindent where the agent's state $\boldsymbol{x}_t = [x_t,y_t,z_t,\theta_t,\phi_t]^\top \in \mathbb{X}$ is composed of position i.e., $\boldsymbol{x}^p_t = [x_t,y_t,z_t]^\top \in \mathbb{R}^3$, and orientation i.e., $\boldsymbol{x}^o_t = [\theta_t,\phi_t]^\top$ components. In particular, $\theta_t \in [-\pi/2,\pi/2]$, and $\phi_t \in [-\pi,\pi]$ denote the vehicle's orientation at time-step $t$ around the $y-$axis, and $z-$axis respectively, as illustrated in Fig. \ref{fig:fig1}. The control input vector $\boldsymbol{u}_t = [u^v_t, u^\theta_t, u^\phi_t]^\top \in \mathbb{U}$ consists of the UAV's linear i.e., $u^v_t$, and angular i.e., $(u^\theta_t, u^\phi_t)$ velocities, $\Delta T$ is the sampling interval, and the normally distributed and uncorrelated random variables $\omega^v_t \sim \mathcal{N}(0,\sigma^2_v)$, $\omega^\theta_t \sim \mathcal{N}(0,\sigma^2_\theta)$, and $\omega^\phi_t \sim \mathcal{N}(0,\sigma^2_\phi)$ model the disturbance on the agent's control inputs. The disturbance vector will be denoted hereafter as $\boldsymbol{\nu}_t = [\omega^v_t ~\omega^\theta_t ~\omega^\phi_t ]^\top \sim \mathcal{N}(\boldsymbol{\bar{\nu}}_t, Q_t)$, where $\boldsymbol{\bar{\nu}}_t = [0~0~0]^\top$, and the covariance matrix $Q_t \in \mathbb{R}^{3 \times 3}$ is given by $Q_t = \text{diag}([\sigma^2_v~\sigma^2_\theta~\sigma^2_\phi])$. Finally, it is assumed that the agent's state is initially distributed as $\boldsymbol{x}_0 \sim \mathcal{N}(\hat{\boldsymbol{x}},\hat{P})$, where $\hat{\boldsymbol{x}}$, and $\hat{P}$ denote the mean and covariance matrix respectively of the agent's state at time-step $t=0$. For brevity, the UAV's dynamical model i.e., Eq. \eqref{eq:dynamics}, will be denoted hereafter as $\boldsymbol{x}_{t+1} = f(\boldsymbol{x}_t,\boldsymbol{u}_t,\boldsymbol{\nu}_t)$, and can potentially be adapted to fit the behaviour of the platform at hand i.e., multi-rotor or fixed-wing UAVs.

\begin{figure}
	\centering
	\includegraphics[width=\columnwidth]{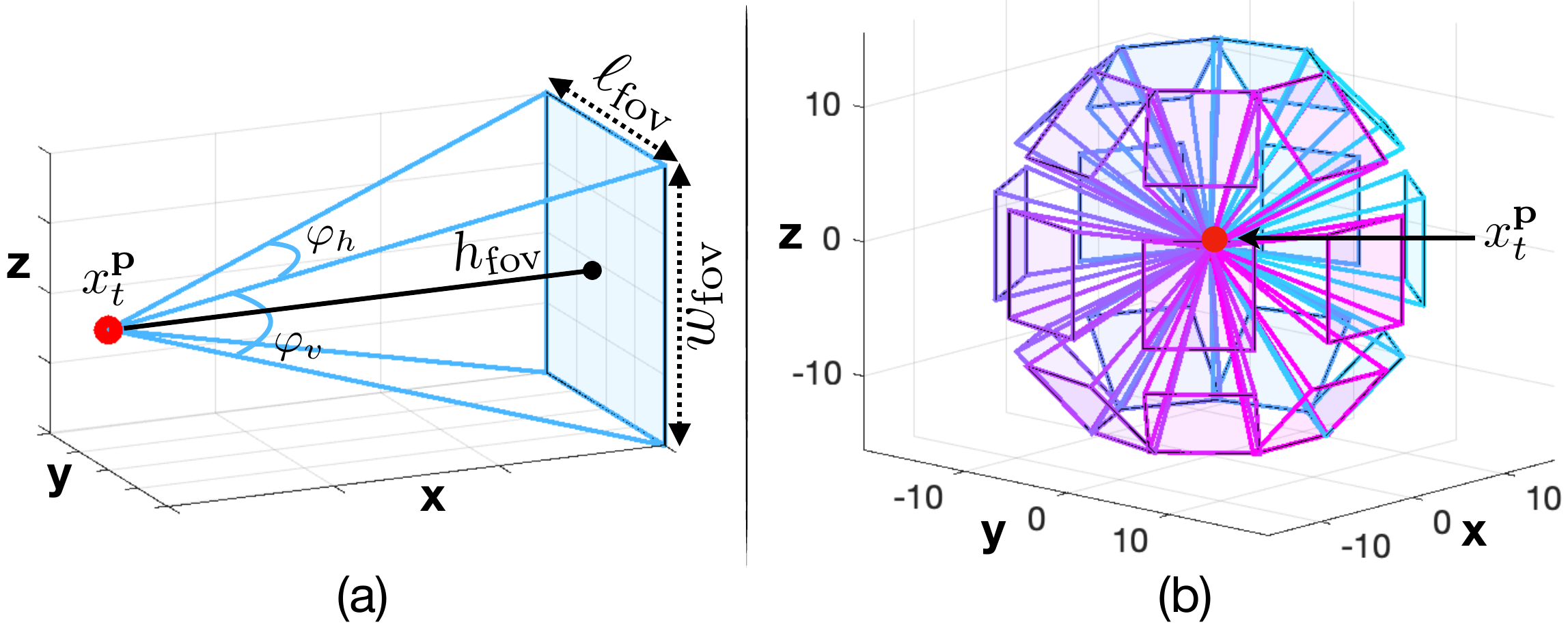}
	\caption{The figure illustrates: (a) the agent's camera FOV characteristics, and (b) the camera FOV rotations $V_m, m \in \{1,..,M\}$ for all admissible pairwise combinations of the rotation angles $(\psi_y,\psi_z) \in \tilde{\Psi}$.}	
	\label{fig:fig2}
\end{figure}

\subsection{UAV Camera Model} \label{ssec:sensing_model}
The UAV agent is equipped with a camera which exhibits a limited field-of-view (FOV), modelled in this work as a regular right pyramid composed of four triangular lateral faces, and a rectangular base. The camera's optical center (given by the agent's position $\mathbf{x}^p_t$) is assumed to be located at the apex which is positioned directly above the centroid of the FOV base. The length $\ell_\text{fov}$, and width $w_\text{fov}$ of the FOV rectangular base (i.e., the size of the projected FOV footprint) is determined by the camera's optical characteristics i.e., the horizontal ($\varphi_h$), and vertical ($\varphi_v$) FOV angles, as $\ell_\text{fov} = 2h_\text{fov} \tan(\frac{\varphi_h}{2})$, and $w_\text{fov} = 2h_\text{fov} \tan(\frac{\varphi_v}{2})$. The parameter $h_\text{fov}$ denotes the camera observation range. The vertices of the camera FOV for a UAV agent positioned at the origin of the 3D cartesian coordinate frame, assuming the camera FOV is pointing forward (i.e., the line, of length $h_\text{fov}$, which is joining the centroid of the base and the apex of the pyramid is parallel to the normal of the $zy-$plane), is given by the 3-by-5 matrix $V_0$ as:
\begin{equation} 
    V_0 =
    \begin{bmatrix}
       h_\text{fov} & h_\text{fov} & h_\text{fov}  & h_\text{fov} &0 \\
       \ell_\text{fov}/2 & \ell_\text{fov}/2 & -\ell_\text{fov}/2 & -\ell_\text{fov}/2 &0 \\
        w_\text{fov}/2  & -w_\text{fov}/2  &  -w_\text{fov}/2  &  w_\text{fov}/2  &0 \\
    \end{bmatrix}.
\end{equation}

The camera's FOV can be rotated in 3D space by executing sequentially two elemental rotations i.e., one rotation by angle $\psi_y \in [-\pi/2,\pi/2]$ around the $y-$axis, followed by a rotation $\psi_z \in (-\pi,\pi]$ around the $z-$axis. Subsequently, at time-step $t$ the pose of the camera FOV can be described by the following geometric transformation:
\begin{equation}\label{eq:fov_vertices1}
    V^i_t(\psi_y,\psi_z) = R_z(\psi_z) R_y(\psi_y) V^i_0 + \boldsymbol{x}^p_t, \forall i \in \{1,..,5\}
\end{equation}

\noindent where $V^i_0$ denotes the $i_\text{th}$ column of the matrix $V_0$, $V^i_t(\psi_y,\psi_z)$ is the corresponding rotated vertex of the FOV parameterized by the rotation angles $\psi_y$ and $\psi_z$, and $\boldsymbol{x}^p_t$ is the UAV's location at time-step $t$. The matrices $R_y(a)$ and $R_z(a)$ represent the basic 3-by-3 rotation matrices \cite{Taubin2011} which rotate a vector by an angle $a$ around the $y-$ and $z-$axis respectively.
%
Finally, it is assumed that the rotation angles $\psi_y \in \Psi_y$, and $\psi_z \in \Psi_z$ take their values from the finite sets of admissible rotation angles $\Psi_y$ and $\Psi_z$ respectively. Let us denote $\tilde{\Psi} = \{(\psi_y,\psi_z) : \psi_y \in \Psi_y, \psi_z \in \Psi_z \}$ the set of all $M = |\tilde{\Psi}|$  pairwise combinations of rotation angles;  there exist $M$ different  orientations of the camera FOV which are given by: $V^i_m = R_z(\psi_z) R_y(\psi_y)V^i_0,~\forall (\psi_y,\psi_z) \in \tilde{\Psi},~ m \in \{1,..,M\},~i\in \{1,..,5\}$, and $V_m \in \mathbb{R}^{3\times5}$ denotes the $m_\text{th}$ orientation, as shown in Fig. \ref{fig:fig2}.



\begin{figure}
	\centering
	\includegraphics[width=\columnwidth]{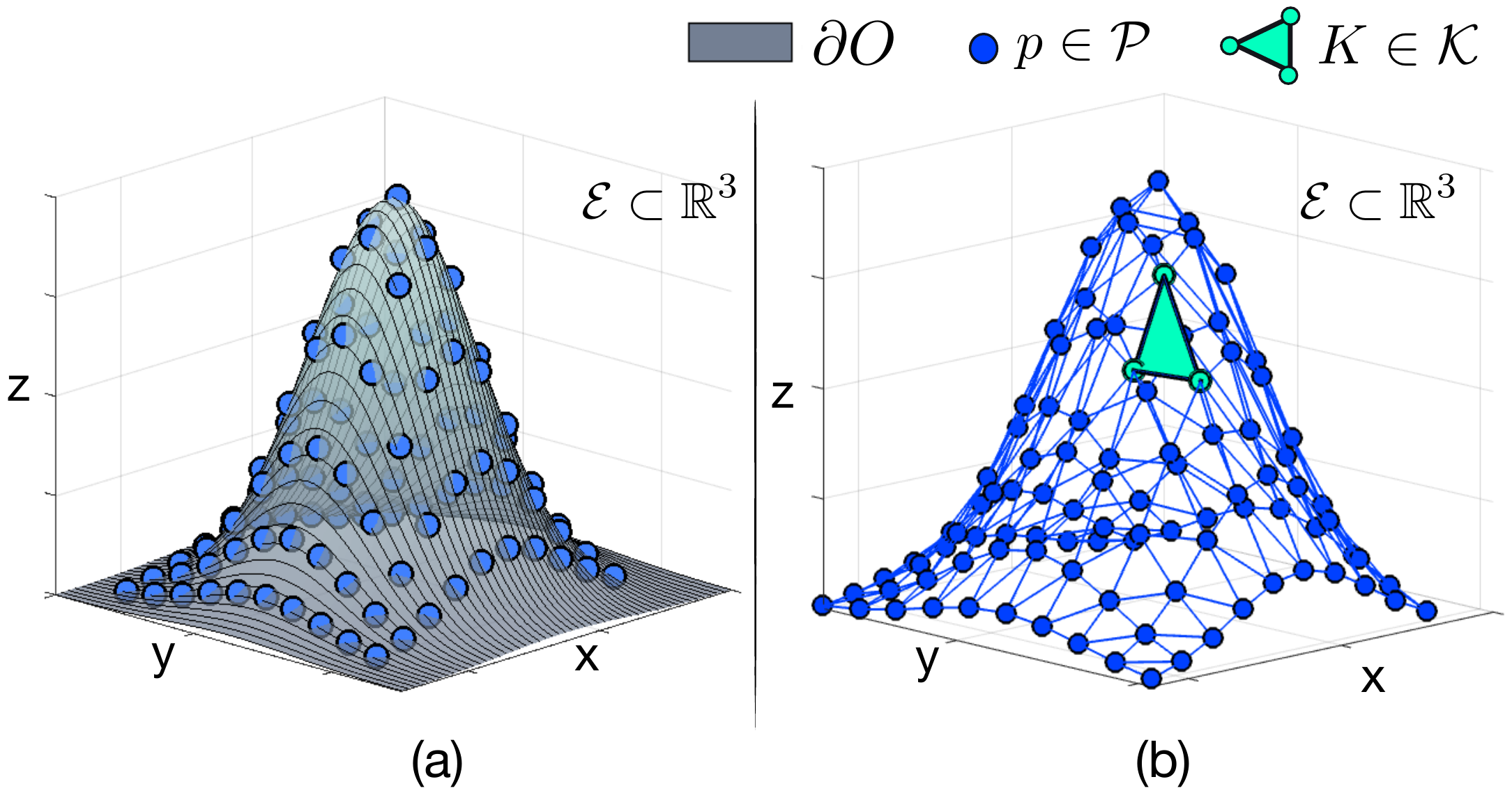}
	\caption{The figure illustrates: (a) The object of interest $\mathcal{O}$ to be covered, and its surface area $\partial \mathcal{O}$ which is represented as a 3D point cloud $\mathcal{P}$, (b) the proposed approach uses Delaunay triangulation to partition the object's surface area into a finite number of non-overlapping triangular facets $K \in \mathcal{K}$, thus forming a 3D mesh of its boundary.}	
	\label{fig:fig3}
\end{figure}

\subsection{Object of interest to be covered} \label{ssec:object_of_interest}
In this work we assume that the surface area $\partial O$ of the convex object of interest $O$ to be covered is represented as a 3D point cloud which was obtained via a 3D scene reconstruction step \cite{Labatut2007}. This 3D point-cloud is then triangulated using Delaunay triangulation \cite{WangT2022} to form a 3D mesh $\mathcal{K}$ which is composed of a finite number of triangular facets $K \in \mathbb{R}^{3\times3} \in \mathcal{K}$, as illustrated in Fig. \ref{fig:fig3}. The objective of the UAV agent is to plan an optimal trajectory which covers a specific subset $\tilde{\mathcal{K}} \subseteq \mathcal{K}$ of the facets on surface area of the object of interest $O$. The number of facets that need to be covered will be denoted hereafter as  $N = |\tilde{\mathcal{K}}|$, where $(|.|)$ is the set cardinality.

\section{Problem Statement}\label{sec:problem}

The problem tackled in this work, can be stated as follows: \textit{Given a sufficiently large planning horizon of length $T$ time-steps, find the UAV's collision-free motion control inputs $\boldsymbol{u}_t,~ t \in \{0,..,T-1\}$, and camera FOV states $V_{{m},t},~ m \in \{1,..,M\}, t \in \{1,..,T\}$, which: a) minimize the mission's cost function $J(\boldsymbol{x}_{1:T},\boldsymbol{u}_{0:T-1})$ that is defined over the UAV control inputs and state trajectory, and b) allow the UAV agent to maximally cover with its camera FOV the subset of facets $\tilde{\mathcal{K}} \subseteq \mathcal{K}$ on the object's surface area.}

During the coverage mission the UAV agent needs to cover all $N$ facets $\tilde{K}_n, ~ n \in \{1,..,N\}$ (where $\tilde{K}_n \in \tilde{\mathcal{K}}$) by appropriately selecting a series of discrete camera FOV states $V_{m,t},~ m \in \{1,..,M\}, t \in \{1,..,T\}$. To achieve this functionality, we follow the procedure outlined next: For each facet $\tilde{K}_n$ on the object's surface that needs to be covered, a waypoint $W_{n} \in \mathbb{R}^{8\times3}$ (represented by the vertices of a cube with length $\ell_W$) is generated. The set of generated waypoints $\mathcal{W}$, are associated with the set of facets $\tilde{\mathcal{K}}$ based on an one-to-one correspondence, and $|\mathcal{W}|=|\mathcal{K}|=N$.

The purpose of the waypoint $W_{n}$ which is associated with the facet $\tilde{K}_n$ is to ensure that when the agent's state $\boldsymbol{x}_{t}$, at some time-step $t$, resides within the convex hull of the waypoint $W_{n}$ (denoted by slight abuse of notation as $\boldsymbol{x}_{t} \in W_{n}$), there exists a camera FOV state $V_{m,t}, m \in \{1,..,M\}$ that results in the coverage of facet $\tilde{K}_n$ i.e., the facet $\tilde{K}_n$ resides within the convex hull of the camera's FOV (with slight abuse of notation this is denoted as $\tilde{K}_n \in V_{m,t}$). Now, because the agent's state is a stochastic quantity, we would like to ensure that each waypoint $W_{n}$ will be visited with a certain probability. This is discussed in more detail in Sec. \ref{sec:approach}. It is important to note here that the optimal waypoint traversal order is determine on the fly i.e., the time-step at which each waypoint is visited is not predetermined, and is decided during optimization. To enable this functionality, along with the selection of the required discrete camera FOV states for coverage, we design and solve an optimal control problem with logical constraints. 
Finally, we would like to make sure that the UAV agent avoids collisions with the obstacles in the environment with certain probability set by the designer. This is discussed in Sec. \ref{ssec:p2_collision}.

\section{Unscented Optimal Control for 3D Coverage Planning}\label{sec:approach}

In this section the details of the proposed coverage planning controller are shown in Problem (P1). Specifically, in Sec.\ref{ssec:p2_ut} we describe on how the unscented transformation is utilized in this work to generate plans based on the future probability distribution of the UAV's state. Then, in Sec.\ref{ssec:p2_guidance} we describe how we have integrated probabilistic and logical constraints into optimal control to enable robust UAV guidance through a series of waypoints, and subsequently in Sec.\ref{ssec:p2_camera} we discuss how coverage is achieved through the control of the UAV's camera. In Sec.\ref{ssec:p2_collision} we present the proposed robust obstacle avoidance technique, and finally the mission's objective cost function is described in Sec. \ref{ssec:p2_objective}.

\subsection{Uncertainty Propagation}\label{ssec:p2_ut}

The constraints shown in Eq. \eqref{eq:P2_1} - Eq. \eqref{eq:P2_12} utilize the unscented transformation (UT) \cite{UT1} in order to estimate the posterior distribution of the state of the UAV agent at each time-step, given the stochastic non-linear motion dynamics shown in Eq. \eqref{eq:dynamics}. 
Specifically, consider the problem of propagating a $d$-dimensional random variable $\bf{x}$, with mean $\bar{\bf{x}}$, and covariance matrix $P_{\bf{x}}$, through the nonlinear function $\bf{y} = f(\bf{x})$. The posterior mean $\bar{\bf{y}}$, and covariance matrix $P_{\bf{y}}$ of the nonlinear transformation $f(.)$, can be computed with the unscented transformation by first generating $2d+1$ sigma points $\mathcal{X}^i,~ i\in\{0,..,2d\}$, and corresponding weights $\mathcal{W}^i,~i\in\{0,..,2d\} $, and then propagating these sigma points through the nonlinearity i.e., $\mathcal{Y}^i = f(\mathcal{X}^i)$. The posterior mean and covariance matrix for $\bf{y}$ are then estimated as the weighted sample mean and covariance of the transformed sigma points.

%
%

Subsequently, the first two constraints in Problem (P1) group together the stochastic terms by concatenating the state and noise random variables, to create a new augmented state i.e., $\boldsymbol{x}^\text{aug}_t \sim \mathcal{N}(\bar{\boldsymbol{x}}^\text{aug}_t,P^\text{aug}_t)$ as shown in Eq. \eqref{eq:P2_1} and Eq. \eqref{eq:P2_2}. The unscented transform is then applied to the new random variable $\boldsymbol{x}^\text{aug}_t \in \mathbb{R}^8$, which has dimension $d=8$, and therefore the constraints in Eq. \eqref{eq:P2_3} - \eqref{eq:P2_8} compute the sigma points $\mathcal{X}^i_t, ~i\in \{0,..,2d\}$ of $\boldsymbol{x}^\text{aug}_t$, and their corresponding weights $\mathcal{W}^i, ~i\in \{0,..,2d\}$. The scaling parameter $\lambda$ is given by $\lambda = \alpha^2(d+\varrho) - d$, where $\alpha$ denotes the spread of the sigma points around the mean, and $\varrho$ is a secondary scaling parameter. The parameter $\beta$ in Eq. \eqref{eq:P2_7} is used for incorporating prior knowledge about the distribution of $\boldsymbol{x}^\text{aug}_t$, and finally, $\left(\sqrt{(d+\lambda)P^\text{aug}_t} \right)_i$ denotes the $i_\text{th}$ row of the square root of the scaled covariance matrix $P^\text{aug}_t$. A detailed discussion on how to choose the parameters of the UT can be found in \cite{UT3}.
The generated sigma points $\mathcal{X}^i_t$ are then propagated through the agent's non-linear dynamics, as shown in Eq. \eqref{eq:P2_9}, for some control input $\boldsymbol{u}_t$ to obtain the transformed sigma points $\mathcal{X}^i_{t+1}$ for the next time-step $t+1$.
These sigma points are then used to approximate the distribution of the agent's state at the next time-step as $\mathcal{N}(\bar{\boldsymbol{x}}_{t+1},P_{t+1})$. This is shown in Eq. \eqref{eq:P2_10} - \eqref{eq:P2_12}.


\subsection{UAV Guidance}\label{ssec:p2_guidance}
\begin{figure}
	\centering
	\includegraphics[width=\columnwidth]{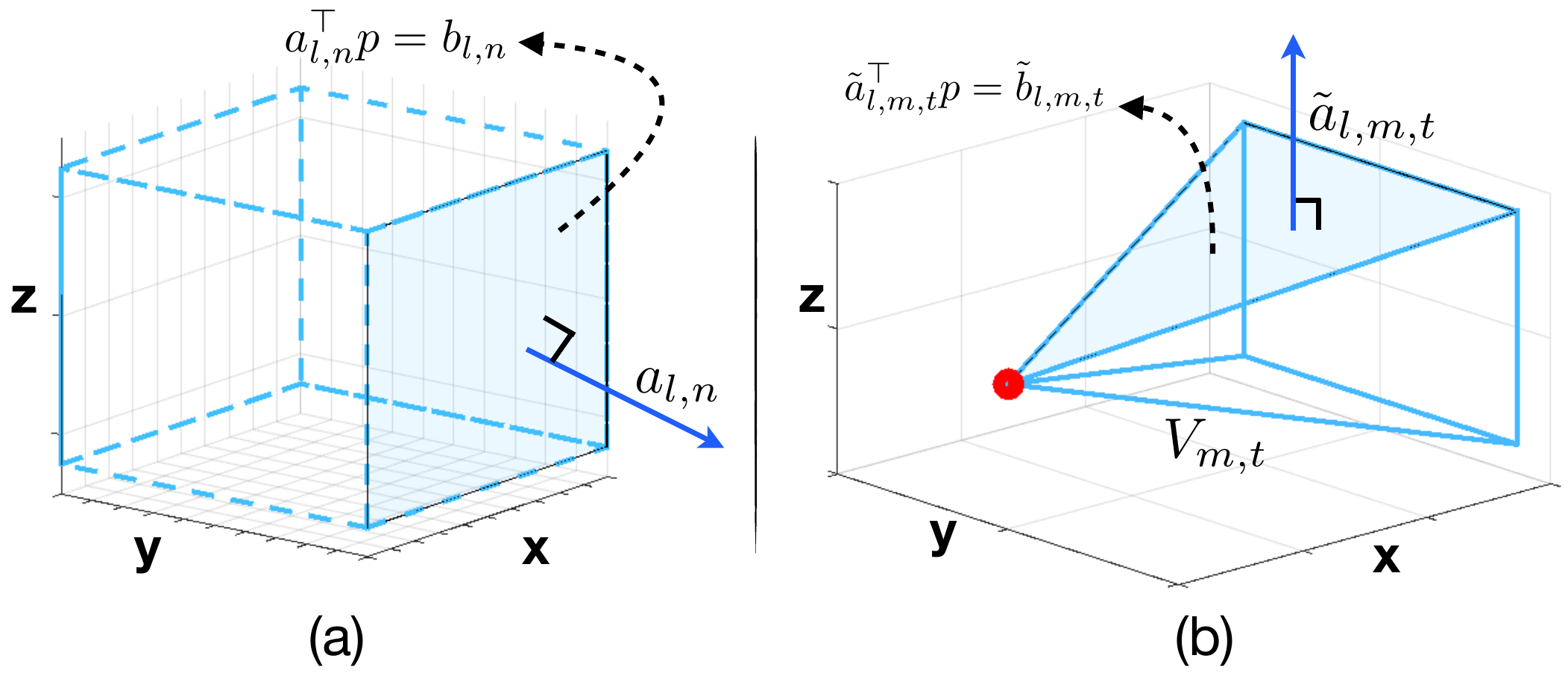}
	\caption{(a) A point $p \in \mathbb{R}^3$ resides within the convex-hull of the cuboid shown in the figure when $a_{l,n}^\top p \leq  b_{l,n},~ \forall l \in \{1,..,6\}$. (b) A point $p \in \mathbb{R}^3$ is covered at time-step $t$ with the agent's $m_\text{th}$ camera FOV $V_{m,t}$ when $\tilde{a}_{\ell,m,t}^\top p \leq \tilde{b}_{\ell,m,t},~ \forall \ell \in \{1,..,5\}$.}	
	\label{fig:fig5}
\end{figure}

\begin{figure}
	\centering
	\includegraphics[width=\columnwidth]{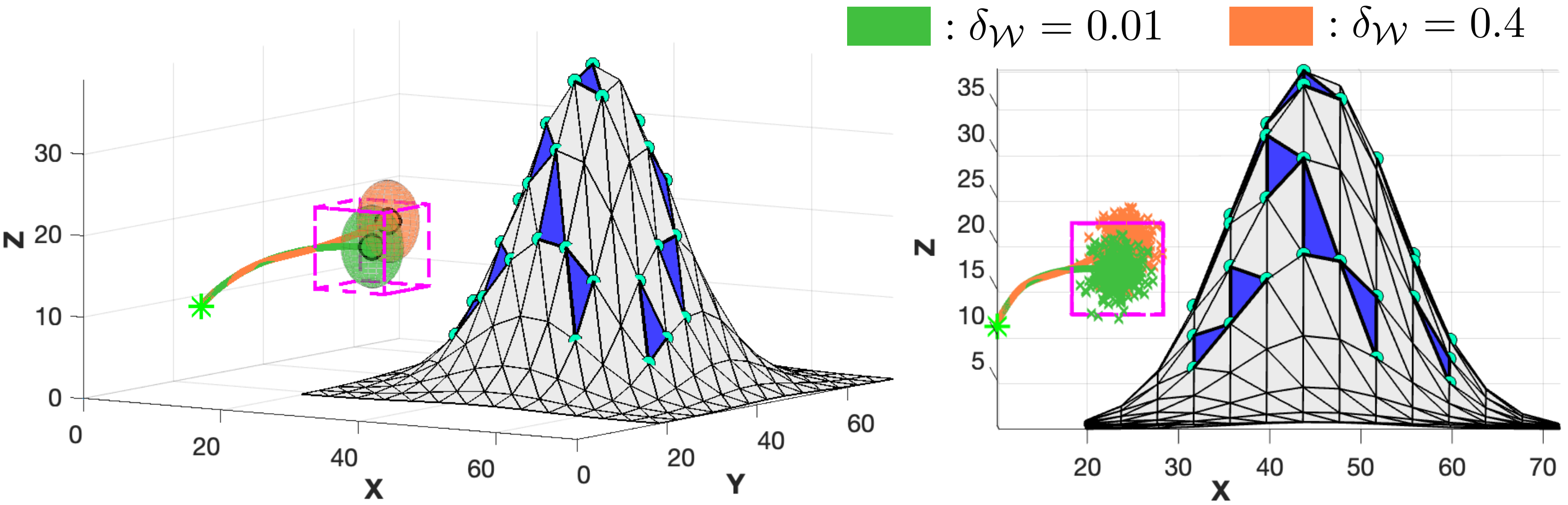}
	\caption{Left: The figure illustrates the UAV's planned trajectory for passing through a single waypoint for two different values of $\delta_\mathcal{W}$. The uncertainty on the UAV state (i.e., the covariance matrix associated with posterior distribution) is depicted in this figure as an error ellipsoid. Right: Various realizations of the UAV's state are drawn from the posterior distribution, and shown as particles for the two values of $\delta_\mathcal{W}$.}	
	\label{fig:fig6}
\end{figure}

Based on the posterior distribution of the agent's state at each time-step, the constraints shown in  Eq. \eqref{eq:P2_13} - \eqref{eq:P2_16} guide the UAV agent to pass from all waypoints $\mathcal{W}$ which are associated with the set of facets $\tilde{\mathcal{K}}$. 
As mentioned previously, for each facet $\tilde{K}_n, n \in \{1,..,N\}$ that needs to be covered, we generate a waypoint $W_{n}$ with centroid $\hat{W}_{n} \in \mathbb{R}^3$ given by:
\begin{equation} \label{eq:waypoint_centers}
	\hat{W}_{n} = c(h_\text{fov} \gamma_{n}) + \tilde{K}^c_n, ~ n \in \{1,..,N\},
\end{equation}

\noindent where $\gamma_{n}$ is the unit normal vector to the plane which contains facet $\tilde{K}_n$, $h_\text{fov}$ is the range of the camera FOV as described in Sec. \ref{ssec:sensing_model}, $\tilde{K}^c_n$ is the centroid of facet $\tilde{K}_n$, and finally $c \in [0,1]$ is a scaling parameter which adjusts the distance between $\hat{W}_{n}$ and $\tilde{K}^c_n$. The generated waypoint $W_n$ is represented in this work by a cube, with length $\ell_W$, centered around the point $\hat{W}_{n}$, and thus $W_n$ is composed of $L=6$ equal faces. The equation of the plane which contains the $l_\text{th}$ face of the $n_\text{th}$ waypoint is given by:

\begin{equation}\label{eq:plane_eq}
  a_{l,n}^\top p =  b_{l,n}, ~ l \in \{1,..,L\}, n \in \{1,..,N\},
\end{equation}

\noindent where $a_{l,n}^\top p$ denotes the dot product between the outward unit normal vector $a_{l,n}$ to the plane which contains face $l$, and the point $p \in \mathbb{R}^3$. $b_{l,n}$ is a constant obtained from the dot product of $a_{l,n}$ with a known point on the plane. Subsequently, a point $p \in \mathbb{R}^3$ which resides inside the convex hull of some waypoint $W_n$, denoted hereafter as $p \in W_n$, must satisfy the following condition:
\begin{equation}\label{eq:inside_cuboid}
  p \in W_n \iff a_{l,n}^\top p \leq  b_{l,n},~ \forall l \in \{1,..,L\}.
\end{equation}

\noindent as illustrated in Fig. \ref{fig:fig5}(a). Therefore, in a deterministic scenario it would suffice to substitute the agent's position i.e., $\boldsymbol{x}_t^p$ for $p$ in Eq. \eqref{eq:inside_cuboid} to determine whether the agent passes through the waypoint $W_n$ at some time-step $t$. However, in the application scenario investigated in this work the agent's state, and therefore its position, is a random variable and as a consequence Eq. \eqref{eq:inside_cuboid} cannot be used directly. 
For notational convenience, let us denote the agent's position at time-step $t$ as $\boldsymbol{x}^p_t \sim \mathcal{N}(\bar{\boldsymbol{x}}^p_t, P^p_t)$, which is extracted from the agent's state at time-step $t$. Then, the constraint shown in Eq. \eqref{eq:inside_cuboid} can be transformed into the following probabilistic form:
\begin{equation}\label{eq:inside_cuboid_prob}
Pr(-a_{l,n}^\top \boldsymbol{x}^p_t + b_{l,n} \leq 0) \leq \delta_\mathcal{W}, ~\forall l \in \{1,..,L\},
\end{equation}
\noindent where $\delta_\mathcal{W}$ is the probability for which the $l_\text{th}$ constraint is not satisfied i.e., $a_{l,n}^\top \boldsymbol{x}^p_t > b_{l,n}$, with probability $\delta_\mathcal{W}$. We can now convert the probabilistic constraint shown in Eq. \eqref{eq:inside_cuboid_prob} into a linear deterministic constraint which takes into account the posterior distribution of the agent's state as follows: First, we define the random variable $y_{l,n} = -a_{l,n}^\top \boldsymbol{x}^p_t + b_{l,n}$ which is distributed according to $y_{l,n}~\sim~\mathcal{N}(-a_{l,n}^\top\bar{\boldsymbol{x}}^p_t+b_{l,n},a_{l,n}P^p_ta_{l,n}^\top)$, and so we need to compute:
\begin{equation} \label{eq:y2}
	Pr(y_{l,n}~\leq~0)~\leq~\delta_\mathcal{W}.
\end{equation}
\noindent We then use the inverse cumulative distribution function of the normal distribution (i.e., the quantile function) to find a deterministic constraint on the mean of $y_{l,n}$ which satisfies Eq. \eqref{eq:y2} for the desired probability $\delta_\mathcal{W}$. More specifically, the quantile function $\Phi^{-1}(\delta)$ of a normal random variable with mean $\mu$ and variance $\sigma^2$ is given by \cite{StatisticalModelling}:
\begin{equation}\label{eq:quantile}
	\Phi^{-1}(\delta) = \mu + \sigma \sqrt{2} \cdot \text{erf}^{-1}(2\delta-1),~ \delta \in [0,1],
\end{equation}
\noindent where $\text{erf}^{-1}(.)$ is the inverse error function. Now we can find the value of $\mu$ which satisfies Eq. \eqref{eq:y2}, by substituting in Eq. \eqref{eq:quantile} $\delta_\mathcal{W}$ for $\delta$, the standard deviation of $y_{l,n}$ for $\sigma$, and then solve for $\mu$ which for $y_{l,n}$ is equal to $\mu = -a_{l,n}^\top\bar{\boldsymbol{x}}^p_t+b_{l,n}$. Therefore we have that:
\begin{align}
\!	&Pr(y_{l,n}~\leq~0)\leq\delta_\mathcal{W} \implies a_{l,n}^\top \bar{\boldsymbol{x}}^p_t - b_{l,n} \leq -\zeta^\mathcal{W}_{l,n,t},
\end{align}
\noindent where $\zeta^\mathcal{W}_{l,n,t} = \sqrt{2 a_{l,n} P^p_t a_{l,n}^\top} \cdot \text{erf}^{-1}(1-2\delta_\mathcal{W})$. The discussion above is demonstrated in Fig. \ref{fig:fig6}.

Based on the  above discussion, the constraint shown in Eq. \eqref{eq:P2_13} checks whether the  expected value of the agent's location i.e., $\bar{\boldsymbol{x}}^\text{p}_{t} \in \mathbb{R}^3$, at time-step $t \in \{1,..,T\}$ satisfies the inequality $a_{l,n}^\top \bar{\boldsymbol{x}}^{\text{p}}_{t} \leq b_{l,n}-\zeta^\mathcal{W}_{l,n,t}$, for the $l_\text{th}$ face of the $n_\text{th}$ waypoint, using the decision variable $w^1_{l,n,t} \in [0,1]$. When the agent's position resides within the $n_\text{th}$ waypoint $W_n$, at some time-step $t$, then $w^1_{l,n,t} > 0, \forall l \in \{1,..,L\}$. On the other hand, when the inequality $a_{l,n}^\top \bar{\boldsymbol{x}}^\text{p}_{t} \leq b_{l,n}-\zeta^\mathcal{W}_{l,n,t}$ is violated, the decision variable $w^1_{l,n,t}$ takes the zero value to satisfy the constraint shown in  Eq. \eqref{eq:P2_13}. Then, the decision variable $w^2_{n,t} \in [-L,0]$ shown in Eq. \eqref{eq:P2_14} is maximized i.e., $w^2_{n,t}=0$ when the agent resides within the waypoint $W_n$ at some time-step $t \in [1,..,T]$ and $w^1_{l,n,t}=1,~ \forall l$. To ensure that $w^1_{l,n,t}=1,~ \forall l$, for some time-step $t$ for waypoint $W_n$, we utilize the constraints shown in Eq. \eqref{eq:P2_15} and Eq. \eqref{eq:P2_16} with the use of the decision variable $w^3_{n,t} \in [0,1]$ which is enforced to take non-zero value for at least one time-step $t \in [1,..,T]$ during the planning horizon for each waypoint. Subsequently, at the time-step $t$ for which $w^3_{n,t} \neq 0$, the constraint in Eq. \eqref{eq:P2_15} forces the decision variable $w^2_{n,t}$ to take a non-negative value, which only occurs when the decision variable $w^1_{l,n,t} = 1, \forall l \in \{1,..,L\}$, which in turn drives the agent's distribution inside the waypoint $W_n$ at time-step $t$ through the constraint in Eq. \eqref{eq:P2_13}.

To summarize, the logical constraints shown Eq. \eqref{eq:P2_13} -  Eq. \eqref{eq:P2_16} which are realized through the continuous decision variables $w^1, w^2$, and $w^3$, allow the UAV agent to visit all $N$ waypoints at least once during the planning horizon by deciding on the fly the waypoint traversal order.

\setcounter{equation}{15}
\begin{algorithm}
\begin{subequations}
\begin{align}
&\hspace{-2mm} \textbf{Problem (P1):}~ \textit{Proposed Unscented 3D} &  \nonumber\\
&\hspace{-2mm} ~~~~~~~~~~~~~~~~~~\textit{Coverage Controller} &  \nonumber\\
&~~\underset{u_0,...,u_{T-1}}{\arg \min} ~\mathbb{E}\left[J(\boldsymbol{x}_{1:T},\boldsymbol{u}_{0:T-1})\right] & \label{eq:objective_P2} \\
&\hspace{-2mm} \textbf{subject to:} ~ t \in \{0,..,T-1\}  &\nonumber\\
&\hspace{-2mm}\texttt{\% Construct augmented state}  &\hspace{-10mm}\nonumber\\
&\hspace{-2mm} \bar{\boldsymbol{x}}^\text{aug}_{t} = [\bar{\boldsymbol{x}}^\top_t~\bar{\boldsymbol{\nu}}^\top_t]^\top,~ \bar{\boldsymbol{x}}_0=\hat{\boldsymbol{x}},  & \hspace{-10mm} \forall t \label{eq:P2_1}\\
&\hspace{-2mm} P^\text{aug}_{t} = \text{blkdiag}([P_t~Q_t]),~ P_0 = \hat{P}, & \hspace{-10mm} \forall t \label{eq:P2_2}\\
&\hspace{-2mm}\texttt{\% Compute sigma points}  &\hspace{-10mm}\nonumber\\
&\hspace{-2mm} \mathcal{X}^0_{t} = \bar{\boldsymbol{x}}^\text{aug}_{t}, & \hspace{-10mm} \forall t \label{eq:P2_3}\\
&\hspace{-2mm} \mathcal{X}^{\tilde{i}}_{t} = \bar{\boldsymbol{x}}^\text{aug}_{t} + \left(\sqrt{(d+\lambda)P^\text{aug}_{t}} \right)_{\tilde{i}} & \hspace{-10mm}  \forall \tilde{i}, t \label{eq:P2_4}\\
&\hspace{-2mm} \mathcal{X}^{\hat{i}}_{t} = \bar{\boldsymbol{x}}^\text{aug}_{t} - \left(\sqrt{(d+\lambda)P^\text{aug}_{t}} \right)_{\hat{i}-d} & \hspace{-10mm}  \forall \hat{i}, t \label{eq:P2_5}\\
&\hspace{-2mm}\texttt{\% Compute sigma weights}  &\hspace{-10mm}\nonumber\\
&\hspace{-2mm} \mathcal{W}_{m^\prime}^{0} = \lambda/(d+\lambda) & \hspace{-10mm} \label{eq:P2_6}\\
&\hspace{-2mm} \mathcal{W}_{c}^{0} = \lambda/(d+\lambda) + (1-\alpha^2+\beta) & \hspace{-10mm} \label{eq:P2_7}\\
&\hspace{-2mm} \mathcal{W}^{\bar{i}}_{c}=\mathcal{W}^{\bar{i}}_{m^\prime} = 1/(2d+2\lambda), & \hspace{-10mm} \bar{i} \in \{1,..,2d\} \label{eq:P2_8}\\
&\hspace{-2mm}\texttt{\% Propagate sigma points}  &\hspace{-10mm}\nonumber\\
&\hspace{-2mm} \mathcal{X}^i_{t+1} = f(\mathcal{X}^i_{t},\boldsymbol{u}_t), & \hspace{-10mm}  \forall i, t \label{eq:P2_9}\\
&\hspace{-2mm}\texttt{\% Compute posterior density}  &\hspace{-10mm}\nonumber\\
&\hspace{-2mm} \bar{\boldsymbol{x}}_{t+1} = \sum_{i=0}^{2d} \mathcal{W}^i_{m^\prime} \mathcal{X}^i_{t+1},  & \hspace{-10mm}  \forall t \label{eq:P2_10}\\
&\hspace{-2mm} \mathcal{Y}^i_{t+1} = (\mathcal{X}^i_{t+1} - \bar{\boldsymbol{x}}_{t+1} ), & \hspace{-10mm}   \forall i, t\label{eq:P2_11}\\
&\hspace{-2mm} P_{t+1} = \sum_{i=0}^{2d} \mathcal{W}^i_{c} \left[\mathcal{Y}^i_{t+1}(\mathcal{Y}^i_{t+1})^\top\right], & \hspace{-10mm} \forall t \label{eq:P2_12}\\
&\hspace{-2mm}\texttt{\% Guidance control}  &\hspace{-10mm}\nonumber\\
&\hspace{-2mm} w^1_{l,n,t+1}\!\! \left[a_{l,n}^\top \bar{\boldsymbol{x}}^\text{p}_{t+1} \! -  \!b_{l,n} \right]\! \leq \! -\zeta^\mathcal{W}_{l,n,t+1} w^1_{l,n,t+1}&\hspace{-10mm}\forall l, n, t\label{eq:P2_13}\\
&\hspace{-2mm} w^2_{n,t+1} = \sum_{l=1}^{L} w^1_{l,n,t+1} - L,  &\hspace{-10mm}\forall n, t\label{eq:P2_14}\\
&\hspace{-2mm} w^3_{n,t+1}w^2_{n,t+1} \geq 0, &\hspace{-10mm}\forall n, t\label{eq:P2_15}\\
&\hspace{-2mm} \sum_{t=0}^{T-1} w^3_{n,t+1} = 1, &\hspace{-10mm}\forall n\label{eq:P2_16}\\
&\hspace{-2mm}\texttt{\% Camera control}  &\hspace{-10mm}\nonumber\\
&\hspace{-2mm} V_{m,t+1} = V_m + \bar{\boldsymbol{x}}^\text{p}_{t+1}, &\hspace{-10mm} \forall m, t\label{eq:P2_17}\\
&\hspace{-2mm} g^1_{\ell,m,n,t+1} \left[\tilde{a}_{\ell,m,t+1}^\top \tilde{K}^c_n - \tilde{b}_{\ell,m,t+1} \right] \leq 0, &\hspace{-10mm}\forall \ell, m, n, t\label{eq:P2_18}\\
&\hspace{-2mm} g^2_{m,n,t+1} = \sum_{\ell=1}^{\tilde{L}} g^1_{\ell,m,n,t+1} -\tilde{L},  &\hspace{-10mm}\forall  m, n, t\label{eq:P2_19}\\
&\hspace{-2mm} g^2_{m,n,t+1} w^3_{n,t+1} s^\text{FOV}_{m,t+1} \geq 0, &\hspace{-12mm} \forall  m, n, t\label{eq:P2_20}\\
&\hspace{-2mm} \sum_{m=1}^{M} s^\text{FOV}_{m,t+1} = 1, &\hspace{-10mm}  \forall t\label{eq:P2_21}\\
&\hspace{-2mm} \max_{}(s^\text{FOV}_{1:M,t+1}) - \min_{}(s^\text{FOV}_{1:M,t+1}) = 1, &\hspace{-10mm} \forall t \label{eq:P2_22}
\end{align}
\end{subequations}
\end{algorithm}

\setcounter{equation}{15}
\begin{algorithm}
\begin{subequations}
\setcounter{equation}{23}
\begin{align}
&\hspace{-8mm} \textbf{Problem (P1):}~ \textit{Proposed Controller}~ \text{(Cont'd)}&  \nonumber\\
&\hspace{-8mm}\texttt{\% Obstacle avoidance}  & \nonumber\\
&\hspace{-8mm} o_{j,\xi,t+1}\left[\hat{a}^\top_{j,\xi}\bar{\boldsymbol{x}}^\text{p}_{t+1}-\hat{b}_{j,\xi}\right] \geq \zeta^\mathcal{O}_{j,\xi,t+1} o_{j,\xi,t+1}, & \hspace{-15mm} \forall j, \xi, t\label{eq:P2_23}\\
&\hspace{-8mm} \sum_{j=1}^{\hat{L}_\xi} o_{j,\xi,t+1} = 1, & \hspace{-15mm} \forall \xi, t\label{eq:P2_24}\\
&\hspace{-8mm}\texttt{\% Operational Bounds}  & \label{eq:P2_25} \\
&\hspace{-8mm} \boldsymbol{x}_t, \bar{\boldsymbol{x}}_t \in \mathbb{X},~ \boldsymbol{u}_t \in \mathbb{U}, &  \hspace{-15mm}\forall t\nonumber\\
&\hspace{-8mm} w^1_{l,n,t},~ w^3_{n,t},~ g^1_{\ell,m,n,t} \in [0,1], & \hspace{-15mm}  \forall l, n, t \nonumber\\
&\hspace{-8mm} s^\text{FOV}_{m,t},~ o_{j,\xi,t} \in [0,1], & \hspace{-15mm} \forall  m, j, \xi, t \nonumber\\
&\hspace{-8mm} w^2_{n,t} \in [-L,0],~ g^2_{m,n,t} \in [-\tilde{L},0], & \hspace{-15mm} \forall  m, n, t \nonumber\\
&\hspace{-8mm} l \in \{1,..,L\},~ \ell \in \{1,..,\tilde{L}\},~ n \in \{1,..,N\}  &\hspace{-15mm}  \nonumber\\
&\hspace{-8mm} m \in \{1,..,M\},~ j \in \{1,..,\hat{L}_\xi\},~ \xi \in \{1,..,\Xi\}  & \hspace{-15mm} \nonumber\\
&\hspace{-8mm} \tilde{i} \in \{1,..,d\}, \hat{i} \in \{d+1,..,2d\}, i \in \{0,..,2d\}   & \hspace{-15mm} \nonumber
\end{align}
\end{subequations}
\vspace{-9mm}
\end{algorithm}

\subsection{UAV Camera Control}\label{ssec:p2_camera}

The constraint in Eq. \eqref{eq:P2_17} computes the $m_\text{th}$ camera FOV state at time-step $t$. As a reminder the camera FOV state $V_{m,t}, ~m \in \{1,..,M\}$ can take at each time step $t$, one out of $M$ different configurations as discussed in Sec. \ref{ssec:sensing_model}. 

A point $p \in \mathbb{R}^3$ resides inside the convex hull of the camera FOV at time-step $t$, when it satisfies the following system of linear inequalities:
\setcounter{equation}{10}
\begin{equation}
	\tilde{a}_{\ell,m,t}^\top p \leq \tilde{b}_{\ell,m,t},~ \forall \ell \in \{1,..,\tilde{L}\},
\end{equation}

\noindent where $\tilde{L} = 5$ denotes the 5 faces of the camera's FOV, $\tilde{a}_{\ell,m,t}^\top p$ is the dot product between the outward normal vector $\tilde{a}_{\ell,m,t}$ on the plane which contains the $\ell_\text{th}$ face of the camera's FOV and the point $p$, and finally $\tilde{b}_{\ell,m,t}$ is a constant, as shown in Fig. \ref{fig:fig5}(b). Therefore, the constraint in Eq. \eqref{eq:P2_18} uses the decision variable $g^1_{\ell,m,n,t} \in [0,1]$ to essentially select the camera FOV state $m$ which at time-step $t$ covers the centroid $\tilde{K}^c_n$ of the $n_\text{th}$ facet $\tilde{K}_n$, which occurs when $g^1_{\ell,m,n,t} >0, \forall \ell$. Next, the variable  $g^2_{m,n,t} \in [-\tilde{L},0]$ takes a non-negative value (i.e., the value of zero) when for some $m$, $n$, and $t$ the decision variable  $g^1_{\ell,m,n,t} = 1, \forall \ell$, as indicated by the constraint shown in Eq. \eqref{eq:P2_19}. The next constraint shown in  Eq. \eqref{eq:P2_20} makes sure that $g^2_{m,n,t}$ will take a non-negative value for at least one FOV state $m$, during the planning horizon for facet $\tilde{K}_n$, which in turn will force the decision variable $g^1_{\ell,m,n,t}$ to take its maximum value for all faces of the $m_\text{th}$ FOV i.e., $g^1_{\ell,m,n,t} = 1, \forall \ell \in \{1,..,\tilde{L}\}$, thus selecting the FOV state which covers the centroid of the $n_\text{th}$ facet. In order to achieve this, we use the decision variables $w^3_{n,t}$, and $s^\text{FOV}_{m,t}$. As we have already discussed, $w^3_{n,t}$ takes a positive value for at least one time-step during the planning horizon for waypoint $W_n$, which is associated with facet $\tilde{K}_n$, i.e., the time when the agent passes through $W_n$. On the other hand, the decision variable $s^\text{FOV}_{m,t} \in [0,1]$ uses constraints shown in Eq. \eqref{eq:P2_21}-\eqref{eq:P2_22} to activate at each time-step $t$ only one out of the $M$ possible FOV states. Therefore, the constraint in Eq. \eqref{eq:P2_20} drives the selection of the FOV state $m$ which covers facet $\tilde{K}_n$ at time-step $t$, by enforcing a non-negative value for the decision variable $g^2_{m,n,t}$ through the decision variables $w^3_{n,t}$, and $s^\text{FOV}_{m,t}$ as shown. We should mention here, that for brevity we have only discussed the constraints which result in the coverage of the centroid point of each facet. However, the constraints in Eq. \eqref{eq:P2_17}-\eqref{eq:P2_22} can be easily extended to cover the 3 vertices of each facet instead of the centroid.

\subsection{UAV Obstacle Avoidance} \label{ssec:p2_collision}
The constraints shown in Eq. \eqref{eq:P2_23} - \eqref{eq:P2_24} make sure that the UAV agent plans a collision-free coverage trajectory by avoiding collisions with the object of interest, and various convex obstacles in the environment. Let us denote the set of all obstacles in the environment (including the object of interest) as $\mathcal{O}$, with cardinality $\Xi = |\mathcal{O}|$. In a deterministic scenario collision avoidance with some obstacle $O_\xi \in \mathcal{O}, ~ \xi \in\{1,..,\Xi\}$, can be achieved by restricting the agent's position $\boldsymbol{x}^{\text{p}}_{t}$ to reside outside the convex-hull of $O_\xi$ for all time-steps $t \in \{1,..,T\}$ during the mission. This is denoted hereafter as: $\boldsymbol{x}^{\text{p}}_{t} \notin O_\xi, ~\forall t \in \{1,..,T\}, \forall \xi \in\{1,..,\Xi\}$. Let the convex-hull of obstacle $O_\xi$ to be given by the intersection of $\hat{L}_\xi$ half-spaces, where the $j_\text{th}$ half-space is generated by the plane equation:
\begin{equation}
	\hat{a}^\top_{j,\xi} p = \hat{b}_{j,\xi},~ j \in \{1,..,\hat{L}_\xi\},
\end{equation}

\begin{figure*}
	\centering
	\includegraphics[width=\textwidth]{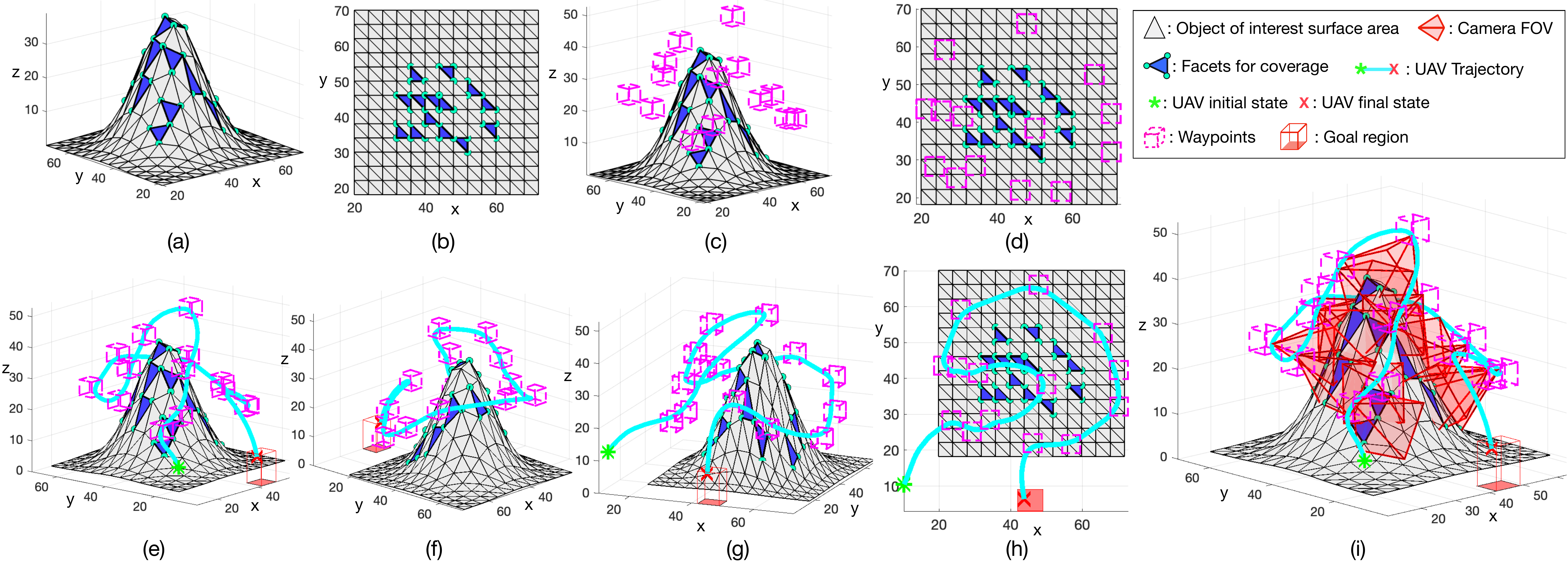}
	\caption{The figure illustrates a simulated 3D coverage scenario with one object of interest. As shown, the UAV is guided through a series of waypoints according to the mission objective, and simultaneously selects the FOV states which result in the coverage of the specified surface area.}	
	\label{fig:res1}
	\vspace{-0mm}
\end{figure*}

\noindent where $\hat{a}_{j,\xi}$ is the unit outward normal vector to the $j_\text{th}$ plane, which generated the $j_\text{th}$ pair of half-spaces, and $\hat{a}^\top_{j,\xi} p$ denotes the dot product of $\hat{a}_{j,\xi}$ with the point $p \in \mathbb{R}^3$. Consequently, a collision is avoided between the agent position $\boldsymbol{x}^{\text{p}}_{t}$ and the obstacle $O_\xi$ at time-step $t$, when the following condition is satisfied: $\exists ~j \in \{1,..,\hat{L}_\xi\} : \hat{a}^\top_{j,\xi} \boldsymbol{x}^{\text{p}}_{t}  \geq \hat{b}_{j,\xi}$. However, because the agent's state is stochastic, the previous constraint cannot be used directly and therefore is transformed into an equivalent probabilistic constraint which can be written as:
\begin{equation} \label{eq:uav_ca_prob}
	Pr(\hat{a}^\top_{j,\xi} \boldsymbol{x}^{\text{p}}_{t} - \hat{b}_{j,\xi} \leq 0) < \delta_\mathcal{O},
\end{equation}

\noindent where $\delta_\mathcal{O}$ is the maximum probability of collision. Following the same procedure described in Sec. \ref{ssec:p2_guidance}, the probabilistic constraint in Eq. \eqref{eq:uav_ca_prob} is converted into a deterministic constraint on the expected value of the UAV's position as shown next:
\begin{equation}\label{eq:uav_ca_prob2}
	\exists ~j \in \{1,..,\hat{L}_\xi\}: \hat{a}^\top_{j,\xi} \bar{\boldsymbol{x}}^{\text{p}}_{t} - \hat{b}_{j,\xi} \geq \zeta^\mathcal{O}_{j,\xi,t},
\end{equation}

\noindent where $\zeta^\mathcal{O}_{j,\xi,t} = \sqrt{2 \hat{a}_{j,\xi} P^p_t \hat{a}_{j,\xi}^\top} \cdot\text{erf}^{-1}(1-2\delta_\mathcal{O})$. Subsequently, obstacle avoidance is enforced with the decision variable $o_{j,\xi,t} \in [0,1]$ as shown in Eq. \eqref{eq:P2_23}-\eqref{eq:P2_24}. Specifically, observe that when $\hat{a}^\top_{j,\xi} \bar{\boldsymbol{x}}^{\text{p}}_{t} < \hat{b}_{j,\xi} + \zeta^\mathcal{O}_{j,\xi,t}$, the decision variable $o_{j,\xi,t}$ becomes $o_{j,\xi,t}=0$ to satisfy the constraint in Eq. \eqref{eq:P2_23}. However, as shown in Eq. \eqref{eq:P2_24}, the variable $o_{j,\xi,t}$ for each time-step $t$, and for each obstacle $O_\xi$ is enforced to take a positive value for some $j \in \{1,..,\hat{L}_\xi\}$. As a result the obstacle avoidance condition in Eq. \eqref{eq:uav_ca_prob2} is enforced for all obstacles at all time-steps with the required probability level.

\subsection{Coverage Mission Objective}\label{ssec:p2_objective}
Finally, the mission objective function i.e., the desired performance index, is defined in this work as the coverage trajectory which utilizes the least amount of energy, and at the same time minimizes the distance between the position of the UAV at the end of the planning horizon with some goal region i.e., to allow the UAV agent to return and land at its home depot at the end of the mission. Thus, Eq. \eqref{eq:objective_P2} is given by:
\begin{equation}\label{eq:cost_function}
	\mathbb{E}\left[J(\boldsymbol{x}_{1:T},\boldsymbol{u}_{0:T-1})\right]  = \sum_{t=0}^{T-1} ||\boldsymbol{u}_t||^2_2 + ||\bar{\boldsymbol{x}}^{\text{p}}_{T} -  \mathcal{G}_o||^2_2,
\end{equation}
\noindent where $\bar{\boldsymbol{x}}^{\text{p}}_{T}$ is the expected value of the agent's position at the end of the planning horizon (i.e., at time-step T), and $\mathcal{G}_o \in \mathbb{R}^3$ is the centroid of the goal region $\mathcal{G}$, which represents the UAV's home depot.

\section{Evaluation} \label{sec:Evaluation}

\subsection{Simulation Setup} \label{ssec:sim_setup}
The evaluation of the proposed approach was conducted through the following simulation setup: The UAV's dynamical model is described by Eq. \eqref{eq:dynamics}, with $\Delta T=0.1$s, linear velocity up to $12$m/s, and angular velocities of up to $\frac{\pi}{3}$rad/s. The disturbance term $\boldsymbol{\nu}_t$ is distributed according to $\boldsymbol{\nu}_t \sim \mathcal{N}(\bar{\boldsymbol{\nu}}_t,Q_t)$ with $\bar{\boldsymbol{\nu}}_t = [0~0~0]^\top$, and $Q_t = 10^{-3}{I}_{3}$, where $I_{3}$ is the $3\times3$ identity matrix. The agent's initial location is distributed according to $\boldsymbol{x}_0 \sim \mathcal{N}(\hat{\boldsymbol{x}},\hat{P})$, with $\hat{\boldsymbol{x}} = [10~10~10~0~0]^\top$, and $\hat{P} = 10^{-4}{I}_{5}$. The UT parameters $\alpha$, $\varrho$, and $\beta$ are set to 1, 2.5, and 2 respectively. The camera FOV is represented as a regular right pyramid composed of $\tilde{L} = 5$ faces, with horizontal and vertical FOV angles set to $\varphi_h = \varphi_v = 60 \deg$, and maximum FOV range $h_\text{fov} = 15$m. We set $\Psi_y = \{-\pi/2, -\pi/4, 0, \pi/4, \pi/2\}$, and $\Psi_z = \{-3\pi/4, -\pi/2,-\pi/4, 0, \pi/4, \pi/2, 3\pi/4, \pi\}$, therefore there are $M=40$ possible camera FOV states. 

The 3D environment $\mathcal{E}$ is bounded in each dimension in the interval $[0,100]$m, and the object of interest $O$ to be covered is given by the Gaussian function $f(x,y) = 40\exp\left(-\left(\frac{(x-45)^2}{160}+\frac{(y-45)^2}{160}\right)\right)$, which has been Delaunay triangulated into $338$ triangular facets $K \in \mathcal{K}$. The set of facets to be covered $\tilde{\mathcal{K}} \subseteq \mathcal{K}$ was randomly sampled from the object's surface area, identifying $N=|\tilde{\mathcal{K}}|=14$ facets $\tilde{K}_n \in \tilde{\mathcal{K}}, n \in \{1,..,N\}$ that need to be covered. For each facet $\tilde{K}_n$, a cubical waypoint $W_n \in \mathcal{W}$ was generated with length $\ell_W = 5$m, $L=6$, and centroid (i.e, center of mass) given by Eq.\eqref{eq:waypoint_centers}, with the scale factor $c$ set to 0.8. The probabilities $\delta_\mathcal{W}$, and $\delta_\mathcal{O}$ have been set to 0.4 and 0.3 respectively. The goal region $\mathcal{G}$ on the ground was represented by a cuboid centered at $\mathcal{G}_o = [45.5~6~5]^\top$. The non-linear program shown in Problem (P1) was solved using off-the-shelf non-linear optimization tools \cite{andersson2019casadi,InteriorPoint}.

\begin{figure}
	\centering
	\includegraphics[width=\columnwidth]{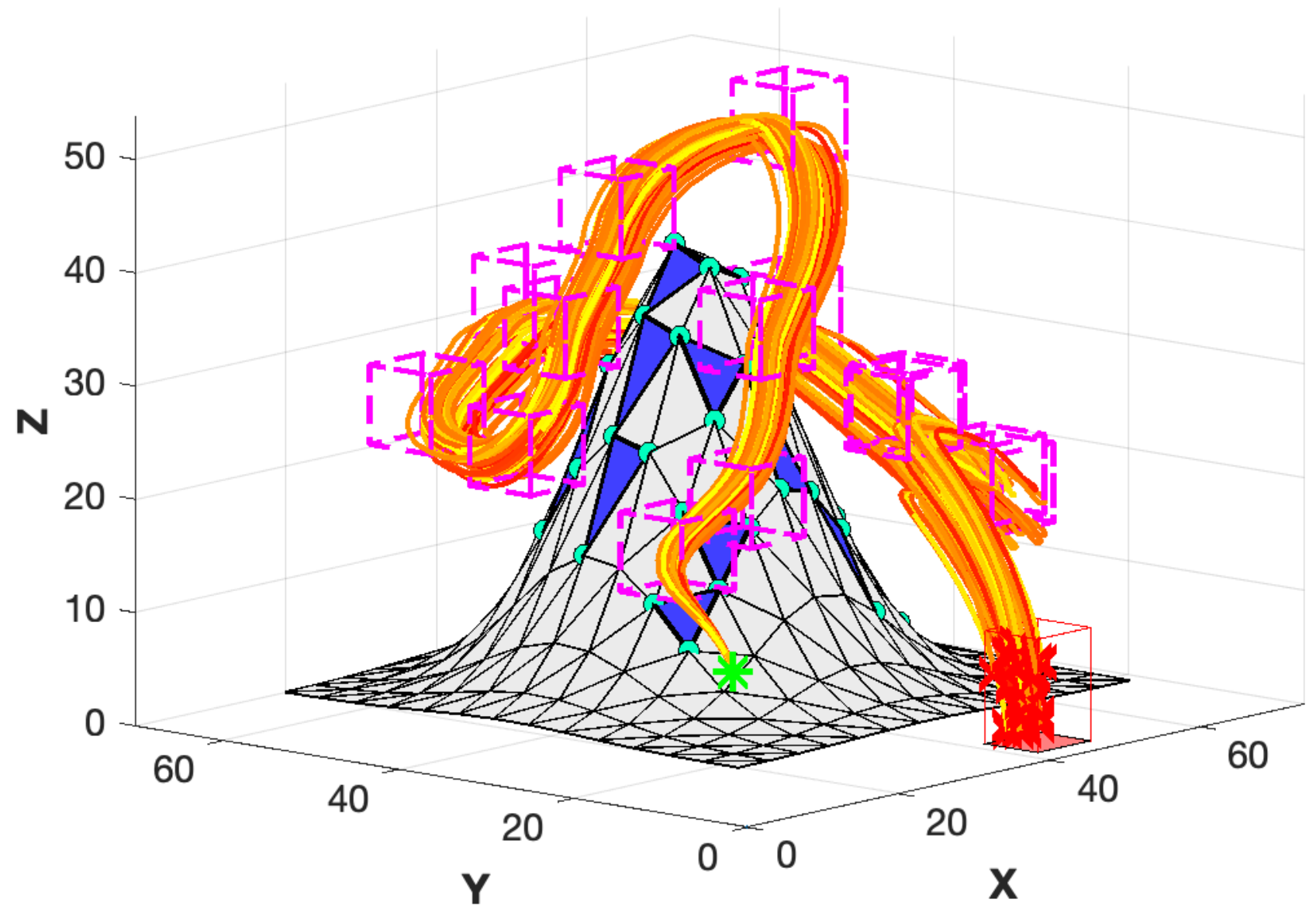}
	\caption{The figure illustrates various realizations of the UAV's coverage trajectory sampled from the posterior distribution.}	
	\label{fig:res2}
	\vspace{-0mm}
\end{figure}

\subsection{Results}
Figure \ref{fig:res1} illustrates the simulation setup described in the previous paragraph, and the resulting UAV coverage plan over a planning horizon of length $T=80$ time-steps. In this scenario the objective of the UAV agent is to find the optimal control inputs according to the objective cost function shown in Eq. \eqref{eq:cost_function}, and at the same time cover with its camera all facets $\tilde{K}_n,~ n \in \{1,..,N\}$ on the object's surface. More specifically, Fig. \ref{fig:res1}(a) and Fig. \ref{fig:res1}(b) show the object of interest in 3D, and top-down views respectively, with the facets to be covered marked with blue color. Then, Fig. \ref{fig:res1}(c) and Fig. \ref{fig:res1}(d) show with pink color the generated waypoints $W_n, n \in \{1,..,N\}$ which are used for the UAV guidance according to the constraints shown in Eq. \eqref{eq:P2_13} - \eqref{eq:P2_16}. Once the UAV agent goes through a waypoint $W_n$, the camera control constraints shown Eq. \eqref{eq:P2_17} - \eqref{eq:P2_22} are responsible for selecting the camera FOV state which covers the corresponding facet $\tilde{K}_n$. More specifically, Fig. \ref{fig:res1}(e) - Fig. \ref{fig:res1}(i) show different views of the generated coverage plan, that allowed the agent to pass from all waypoints $\mathcal{W}$, and cover all facets $\tilde{\mathcal{K}}$. In this figure the UAV's trajectory shown in cyan color corresponds to the expected value of the agent's position $\bar{\boldsymbol{x}}^p_{1:T}$ which has been computed with the unscented transformation i.e., Eq. \eqref{eq:P2_1} - \eqref{eq:P2_12}. As we can observe the UAV's expected state is guided through the waypoints, and its final state reaches at the end of the horizon the goal region indicated by the red cuboid.  Finally, Fig. \ref{fig:res1}(i) shows the complete 3D coverage plan, illustrating the selected camera FOV states which have been used to cover all the required facets.

\begin{figure}
	\centering
	\includegraphics[width=\columnwidth]{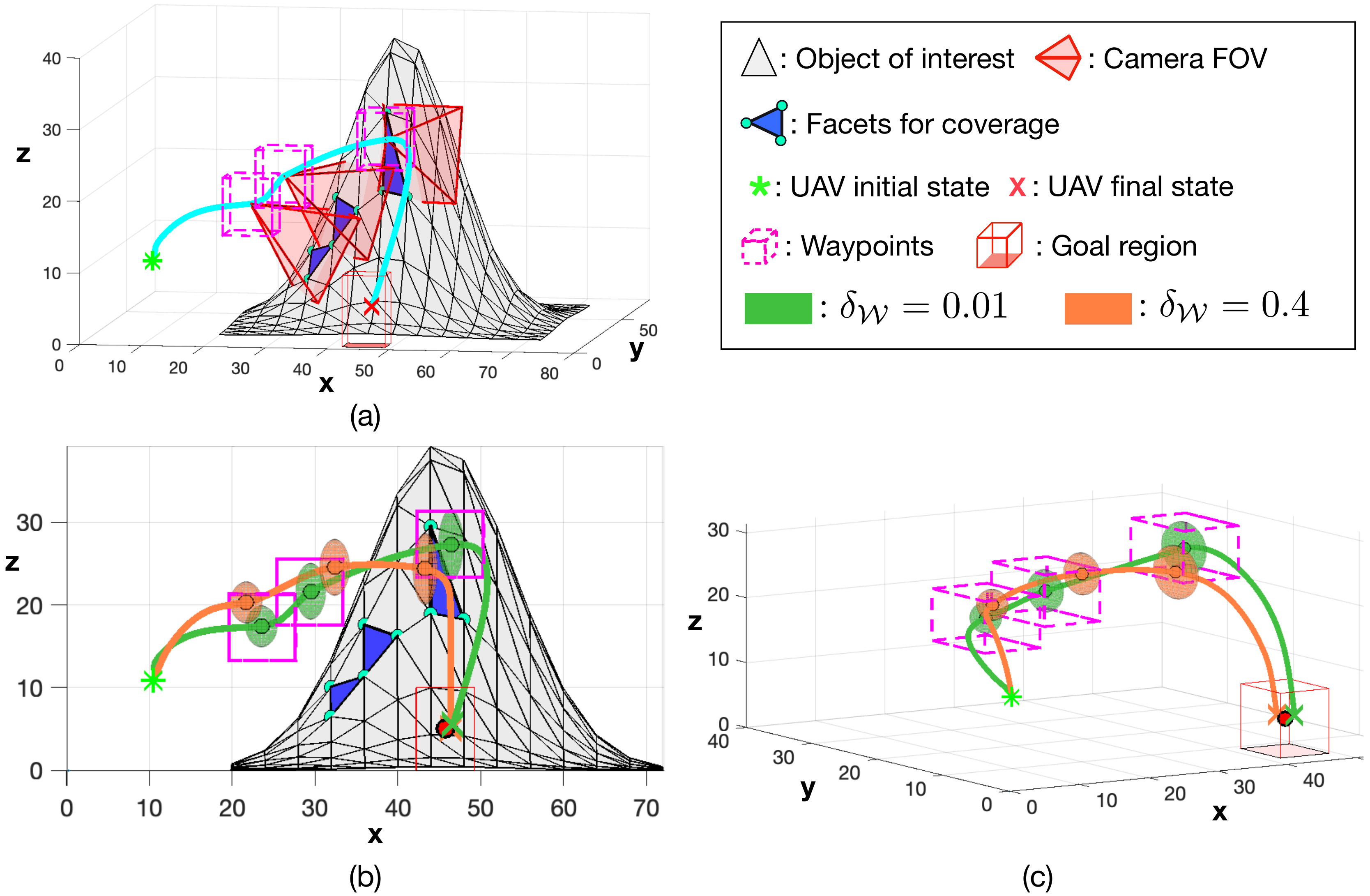}
	\caption{The figure illustrates the robustness of the proposed approach for different levels of the probability $\delta_\mathcal{W}$ used for the UAV guidance through the waypoints.}	
	\label{fig:res3}
	\vspace{-0mm}
\end{figure}

Then, for the same experiment, Fig.\ref{fig:res2} shows 1000 realizations $\boldsymbol{x}^{i}_{1:T}, i=\{1,..,1000\}$ of the UAV's trajectory sampled from the computed posterior distribution of the agent's state i.e., $\boldsymbol{x}^i_{t} \sim \mathcal{N}(\bar{\boldsymbol{x}}_t, P_t), t \in \{1,..,T\}$. The result verifies the applicability of the proposed technique to robustly guide the UAV agent through its coverage mission in the presence of uncertainty. The robustness of the generated coverage trajectories is also demonstrated in Fig. \ref{fig:res3} for two different values of the probability $\delta_\mathcal{W}$. More specifically, Fig. \ref{fig:res3}(a) shows the expected value of the agent's trajectory while passing through 3 waypoints, and the camera FOV states that were selected for the coverage of the 3 facets. This result was obtained with $\delta_\mathcal{W}=0.01$. Then, Fig. \ref{fig:res3}(b)-(c) illustrates the effect of $\delta_\mathcal{W}$ on the generated coverage trajectory. Specifically, the green line indicates the initial trajectory which was obtained with $\delta_\mathcal{W}=0.01$, and the orange line indicates the coverage trajectory which was obtained with $\delta_\mathcal{W}=0.4$. The figure also shows the uncertainty on the position of the UAV at the time-steps for which the UAV passes through the waypoints. This uncertainty is shown here as the 99.9\% confidence error ellipsoid of posterior distribution of the agent's state. When $\delta_\mathcal{W}$ is equal to $0.01$ the error ellipsoid (i.e., denoting the distribution of the agent's position) moves closer the center each waypoint in order to satisfy the probabilistic constraints in Eq. \eqref{eq:y2} at the $0.01$ probability level. However, when $\delta_\mathcal{W}=0.4$, the frequency of cases where the UAV misses the waypoint increases, as indicated by the ellipsoid volume outside the waypoints. 

\begin{figure}
	\centering
	\includegraphics[width=\columnwidth]{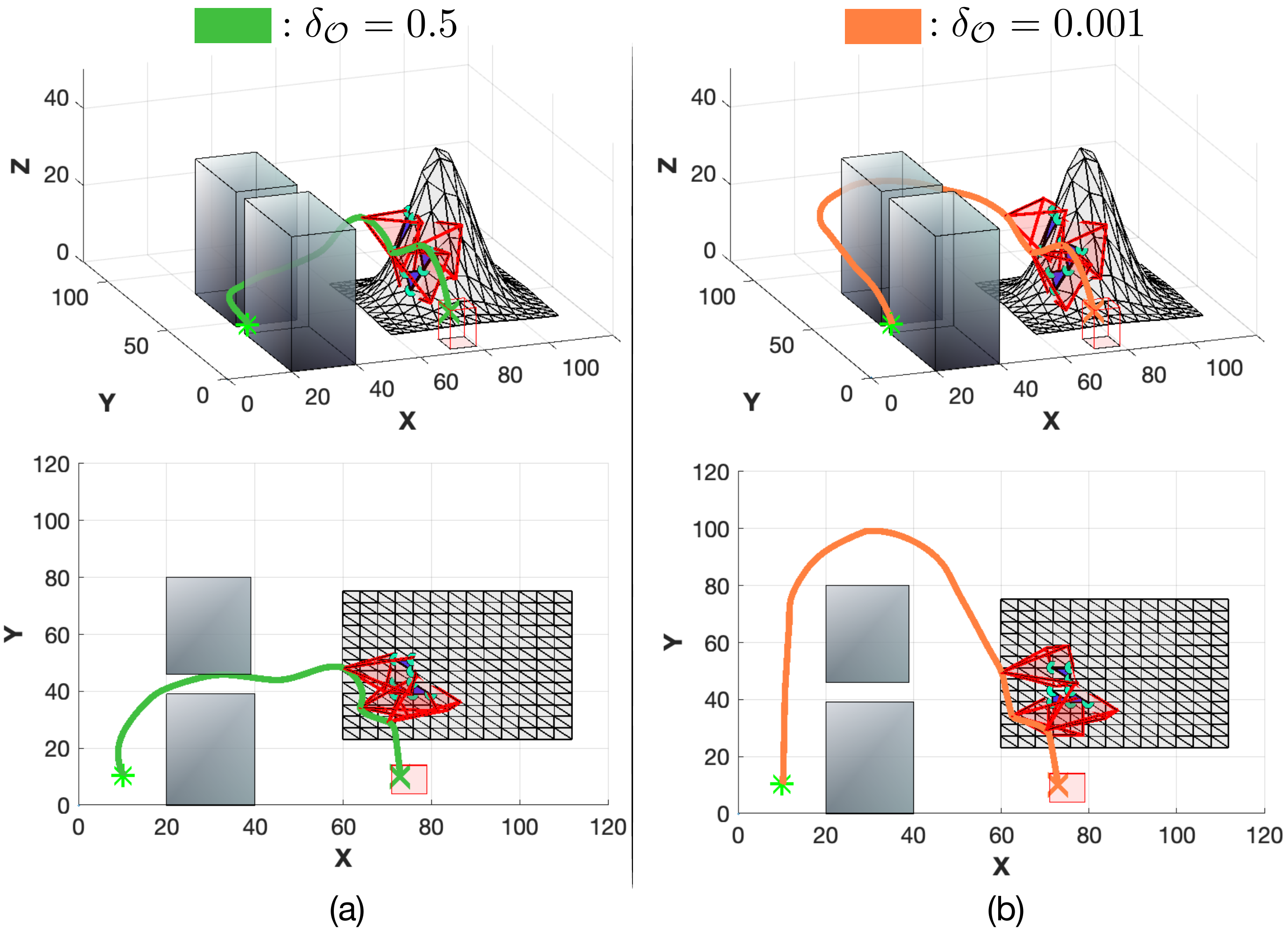}
	\caption{The figure illustrates the behavior of the proposed controller for two different values of the probability of collision $\delta_\mathcal{O}$.}	
	\label{fig:res4}
	\vspace{-3mm}
\end{figure}

Finally, Fig. \ref{fig:res4} illustrates the UAV's behavior for different values of $\delta_\mathcal{O}$ i.e., the probability of collision. In particular, Fig.\ref{fig:res4}(a) shows the UAV's coverage plan for the scenario depicted with the maximum probability of collision set to  $\delta_\mathcal{O}=0.5$, whereas Fig. \ref{fig:res4}(b) shows the same scenario with $\delta_\mathcal{O}=0.01$. As shown in Fig. \ref{fig:res4}(a), when the probability of collision is set to 0.5, the UAV manages to pass through the narrow corridor between the obstacles. However, when the probability of collision is set to less or equal to $0.001$, the planned trajectory can no longer pass between the two obstacles. Instead in order to ensure a safe path at this probability level, the UAV go around the obstacles as shown in Fig. \ref{fig:res4}(b).

\vspace{-3mm}
\section{Conclusion} \label{sec:conclusion}
In this work we have proposed a novel coverage planning approach which utilizes the unscented transformation to compute probabilistically robust open-loop trajectories for 3D object coverage. The proposed optimal control coverage problem integrates logical constraints to allow the joint optimization of the UAV's motion and camera control inputs. We have showed how the logical expressions in the constraints can be realized using equality/inequality constraints involving only continuous variables, and how the probabilistic constraints can be converted into deterministic constraints on the mean state of the agent. Simulation results show the effectiveness of the proposed approach. Future work will focus on the implementation and real-world performance of the proposed approach. 

\section*{Acknowledgments}
This work has been supported by the European Union's H2020 research and innovation programme under grant agreement No 739551 (KIOS CoE - TEAMING) and from the Republic of Cyprus through the Deputy Ministry of Research, Innovation and Digital Policy.
\flushbottom
\balance

\bibliographystyle{IEEEtran}
\bibliography{IEEEabrv,main} 

\end{document}